\documentclass{article}

\usepackage{microtype}
\usepackage{graphicx}
\usepackage{subfigure}
\usepackage{booktabs} 

\usepackage{hyperref}

\usepackage[accepted]{icml2020}

\usepackage{amsthm}
\theoremstyle{definition}
\newtheorem{definition}{Definition}[section]

\usepackage{amsfonts}

\icmltitlerunning{The Need for Standardised Explainability}

\graphicspath{ {figs/} }

\begin{document}

\twocolumn[
\icmltitle{The Need for Standardised Explainability}

\icmlsetsymbol{equal}{*}

\begin{icmlauthorlist}
\icmlauthor{Othman Benchekroun}{dathena}
\icmlauthor{Adel Rahimi}{dathena}
\icmlauthor{Qini Zhang}{dathena}
\icmlauthor{Tetiana Kodliuk}{dathena}
\end{icmlauthorlist}

\icmlaffiliation{dathena}{Dathena Science Pte. Ltd., Singapore, Singapore}

\icmlcorrespondingauthor{Othman Benchekroun}{othman.benchekroun@dathena.io}
\icmlcorrespondingauthor{Adel Rahimi}{adel.rahimi@dathena.io}
\icmlcorrespondingauthor{Qini Zhang}{qini.zhang@dathena.io}
\icmlcorrespondingauthor{Tetiana Kodliuk}{tania.kodliuk@dathena.io}

\icmlkeywords{XAI, Explainability, Auditability, Interpretability, Black-Box Models, IXAI Models}

\vskip 0.3in
]



\printAffiliationsAndNotice{}  

\begin{abstract}
Explainable AI (XAI) is paramount in industry-grade AI; however existing methods fail to address this necessity, in part due to a lack of standardisation of explainability methods. The purpose of this paper is to offer a perspective on the current state of the area of explainability, and to provide novel definitions for Explainability and Interpretability to begin standardising this area of research. To do so, we provide an overview of the literature on explainability, and of the existing methods that are already implemented. Finally, we offer a tentative taxonomy of the different explainability methods, opening the door to future research.
\end{abstract}

\section{Introduction}
It is undeniable that we are living in the era of Artificial Intelligence (AI). News outlets are talking continuously about an AI revolution, while some public figures such as Andrew Ng went as far as baptizing AI “the new electricity” \cite{ai_electricity}, and with good reason. Today, AI has applications in almost every domain, including ---but not limited to--- Finance, Law Enforcement, Healthcare, Data Privacy \& Security ... Among the best-known applications of the last decade developed in Machine Learning are \textit{Transformer Networks} \cite{vaswani2017attention} (which revolutionized the architecture of Deep Learning NLP networks), \textit{word2vec} \cite{mikolov2013efficient} (which advanced pre-trained word embeddings), and of course \textit{Convolutional Neural Networks} (CNNs) \cite{krizhevsky2012imagenet} (which were used in the popular image classification competition ImageNet \cite{deng2009imagenet} and led to a significant paradigm shift in AI). 

But while such advances dominate the public discourse, dissonant voices have started emerging to mitigate AI’s success. Indeed, there is a major issue with recent machine learning models: although they yield great results, all of them are "black-box models". Their results cannot be interpreted, and their inner-workings cannot be understood. In essence, anything could happen inside the model without us never knowing about it. And there are already many accounts of unaccounted behavior appearing in productized models.

\subsection{Unintentional Misbehavior}

Even though one of AI's biggest pitfalls is to be intentionally misused, it doesn't pose such a great danger as it follows a given expected behavior. The real danger lies in the errors that were not accounted for, and that affect millions of people the same way that AI Weapons do, if not more. These errors, or biases, are inherited from the humans building the AI models and introducing their unconscious systemic discrimination belief system. And if we are not cautious, these biases become self-amplifying.

The latest example of a long streak of biases in technology is the "racist soap dispenser" \cite{racist_soap}. We cannot really blame AI for being inherently racist, it is simply the way that it was programmed and the data on which it was trained that is flawed. When a group of white researchers developed an automatic faucet, they couldn’t predict that it wouldn’t recognize the hands of black people as they only tested it on themselves and on people who looked like them.

The same way, when advertisements for job openings were tailored to a specific audience, there was a noticeable gender bias as the AI only showed top positions to men and almost never to women \cite{sexist_ad}. The most troubling element of this story is that the ad targeting system was developed by Google, and not simply by a small company dabbling with AI. Today, five years after both of these articles, we are finally seeing a slow awakening to these issues.

\subsection{Mistrust in Machine Learning}

Even when the ML algorithms are offering high accuracy, this doesn't mean that the end-user will necessarily trust it and use it as mentioned in \cite{ribeiro2016should}. This problem can be seen as a modern "Occam's Razor" principle where the two conjucts would be the accuracy of a model and its explainability. We can define this paradigm as follows: 

\begin{definition}
Given two models with a similar accuracy ($\pm \varepsilon \%$ with $\varepsilon$ an arbitrary value depending on the problem solved by the AI models), it is always preferable to choose the most explainable model.
\end{definition}

This definition reflects the basis of common ML algorithms. As explained in \cite{goodman2017european}, conventional models learn the correlation between the input data and the output without learning the causality.

\subsection{Contributions of this Paper}

Following the realization that black-box models were dangerous, multiple methods were developed to try to explain them. In this paper, we will first try to give a detailed definition of "Explainable AI" and its main characteristics. We will then go through the main inherently explainable models (white-box or clear-box models), before giving an overview of the current methods that can be used to explain and interpret black-box models. To contextualise better our outline, we will provide visual examples aimed at giving more information on the methods presented. Finally, we will point a critical question on why the current methods fail to address industry-grade explainability and suggest a path to standardisation to address these shortcomings.

\section{The need for Explainable AI in the industry}
There are many reasons that drive the need for Explainable AI (XAI) in research, mostly to avoid the shortcomings mentioned above. However, there are only two main needs that drive the use of XAI in organizations:
\begin{enumerate}
    \item Organizations \textbf{have to} use explainable AI in order to comply with the privacy-related regulations all over the world putting the data subjects back in control of their personal information. Indeed, if we look at GDPR in Europe, at PDPA in Singapore, at CCPA in California or even at LGPD in Brazil, all these laws have one thing in common: in some way, it forces companies to explain to clients/users the inner workings of their algorithms and how specific decisions have been made.
    \item Organizations \textbf{want to} implement XAI seriously given all the advantages that explainability offers. Not only do organizations gain valuable insights on the behaviour of each of the models they use, they can also build more efficient, profitable and cost-wary AI models. In a global market where consumers are increasingly cautious of how their data is saved and used, actions showing an organization's commitment to transparency through explainability won’t go unnoticed.
\end{enumerate}

\section{Characteristics for Explainable AI}
Now that we have outlined why we need explainable AI, the main questions that remain are: how exactly to define explainable AI, and what should be the characteristics of an explainable model. To answer these questions, we have to take a step back and look at the word \textit{explanation} itself. According to the Merriam-Webster dictionary, \textbf{to explain} is \textit{"to make something plain or understandable"} \cite{explain}. According to this definition, an explainable AI should be understandable by the user, which is the opposite of the so-called "black-box models".

Today's explainability framework uses different words to refer to an Artificial Intelligence that is understandable. “Interpretable”, “Auditable” and “Explainable” are often used interchangeably, but each one represents a different level of understanding of what happens inside an AI model. Because every explanation needs an interrogation, we define three questions below to understand the differences between these three terms and define them properly:

\subsection{Interpretability: “How does the AI model behave?”}

Even though this question might seem straight-forward, it is already the first step in understanding what goes on inside the AI model. Indeed, interpretable AI models make it possible to understand why some data points aren’t well predicted --- and more broadly why the model doesn't behave as expected. There is not yet a formal definition of \textbf{interpretability} in the literature, but researchers have tried to define it as follows in the context of Machine Learning:

\begin{definition}
The interpretability of AI models refers to the condition that allows a user to understand the relationship between the input and the output of the AI models, providing a clear grasp of all resulting data points.
\end{definition}

\subsection{Auditability: “Does the AI model behave as expected?”}

Auditability does not only focus on understanding the current behaviour of the model, but more generally on ensuring that the AI model behaves as expected. It is auditability that allows ensuring that no human biases are introduced in the algorithm being developed, for example through reasoning errors. It is important to note that auditable AI will not eradicate all issues linked to human bias, but it will allow addressing most of them and evaluating the AI models we are building through time to ensure their performance is as expected. 

As the field of AI auditing is still new, there is no real framework or methodology to ensure trust in AI models. However, a definition of auditability would be fruitful. We can thus define the most important property of an auditable AI model as its capacity to answer specific questions auditors might have such as: \textit{“Are there any bugs?”}, \textit{“Are unintentional biases included in the model?”}, \textit{“Does this model have security risks?”} or even \textit{“Is this AI model compliant with a specific data protection regulation?”} \cite{ai_audit}.

\subsection{Explainability: “How is every decision taken by the AI model?”}

Explainability --- the main topic of this paper --- is the most constraining paradigm allowing to understand the mechanics of the AI model precisely. Indeed, to answer this question we need to show the user how the model's parameters are involved in its decision process, and what these parameters represent as a whole. In this sense, we can define explainability as follows:

\begin{definition}
The explainability of AI models refers to the condition that allows a user to understand decisions made by the model and its subparts through processes before, during, and after the construction of the AI model.
\end{definition}

This offers an even more robust assurance than interpretability and auditability that the algorithm behaves as expected as we can dissect every constituting elements and interrogate their choices. Moreover, we provide an all-encompassing answer as we can explain parameters and subparts both separately or together, allowing to give more context to possible unclear or confusing parameters.

\section{Explainability Methods}

After having defined the meaning of \textit{explainability}, the question that needs to be answered is how to actually achieve it and offer a better understanding of what is going on under the hood to the end user. The two main types of explainability presented below are inherent explainability, which consists in using white-box models, and external explainability methods, which are a recourse when it isn't possible to use white-box models.

\subsection{Inherently Explainable Models}
Several AI models are inherently interpretable and explainable. Inherently explainable AI (IXAI) models are white-box models that are transparent by design and highlight the main features used for prediction. Using these allows us to get an intuition of why the model has predicted a certain data point belonging to a certain class by looking at the most important parameters taken into account.

\subsubsection{Decision Trees}

Decision Trees, also known as Classification And Regression Trees (CART), are supervised predictive models that are named after their tree-like structure. Decisions trees are built through recursive binning (which can be binary or multiary) where each condition --- or partition --- is defined as a decision node. Thus, prediction through decision trees can be seen as solving a propositional calculus formula. The main advantage of this architecture is that decision trees are highly interpretable white-boxes.

\begin{figure}[ht]
    \vskip 0.2in
    \begin{center}
    \centerline{\includegraphics[width=\columnwidth]{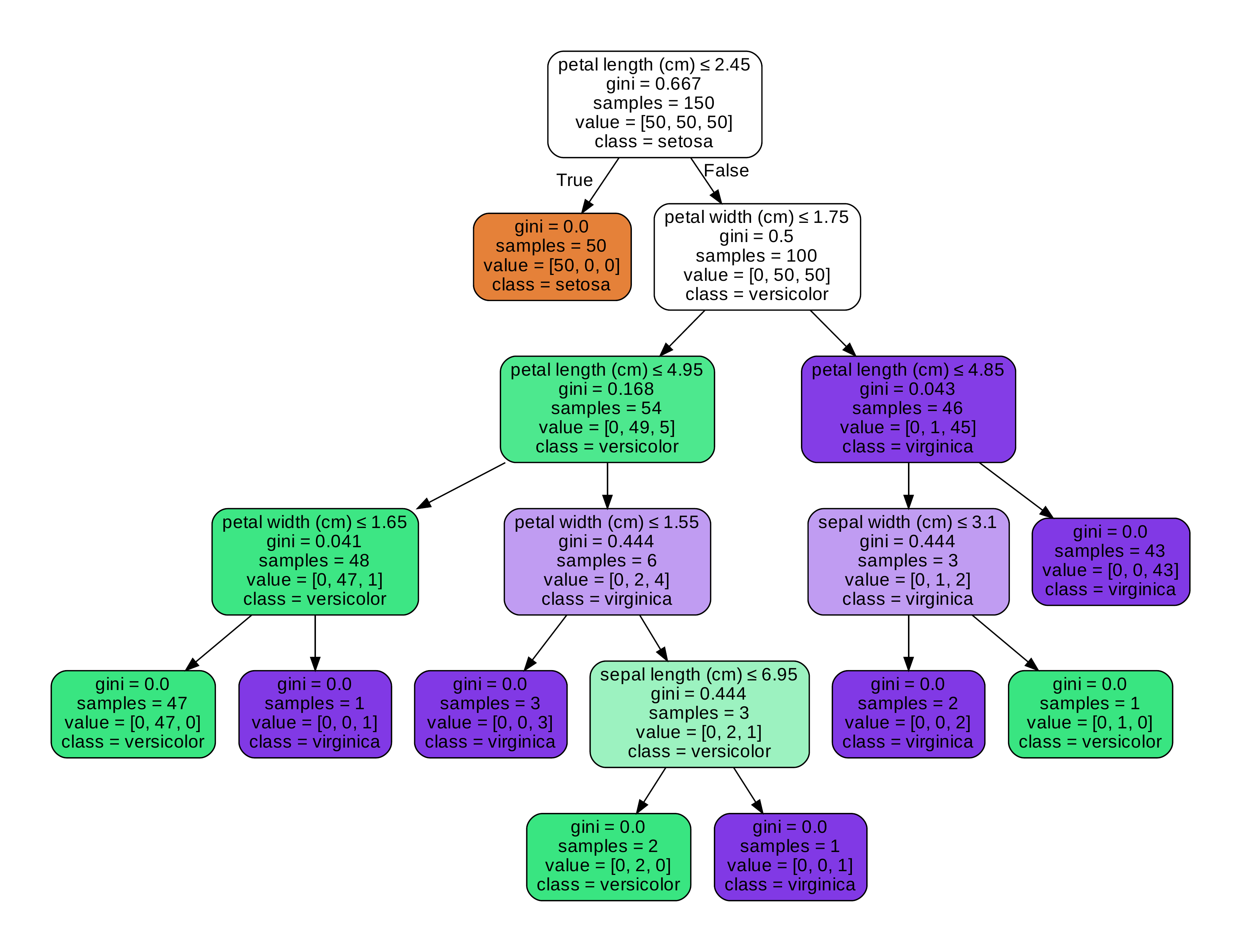}}
    \caption{A plotted Decision Tree with its attributes and labels, trained on Iris dataset \cite{fisher1950use}}
    \label{fig:decision_tree}
    \end{center}
    \vskip -0.2in
\end{figure}

In the figure \ref{fig:decision_tree}, we have plotted a decision tree that was trained on the Iris dataset \cite{fisher1950use}.

Decision trees are very inexpensive and extremely fast to build; however they are not robust, and therefore extremely prone to overfitting. For this reason, we prefer to use Random Forests, which are ensemble methods that build on multiple decision trees to output the final prediction. It follows from this that the results are not inherently explainable anymore, even though it is still possible to understand the decisions taken by all its subparts.

\subsubsection{Linear Predictors}
Much like decision trees, linear predictors are supervised predictive models that can be used either to classify data, or more often to predict continuous values from the input variables. The name "Linear Predictor" comes from the fact that the results are given by computing the linear combinations of all input parameters. The explainability of this model comes from the fact that all parameters have associated weights computed during the training phase. 

Even though linear predictors are prone to overfitting just like decision trees, the solution to this challenge does not bring less explainability but more. Indeed, using regularization to shrink coefficients allows us to determine which factors (or variables) are the most important in our linear combination.

\subsection{Interpreting Black-box Models}

Unfortunately, not all problems are simple enough to be solved by IXAI models and require more complex ML models or Deep Learning (DL) models that aren't so transparent. To provide explainability to these models, multiple methods were created since 2016 to bring understanding to black-box models. Below, we will present chronologically the three methods which have left the biggest impact in the field of XAI.

Some of these methods are based on the idea of "attention", which consists in studying the features which the models "focuses" on (e.g. highlighting the important keywords in a text, or the most important pixels of an image). Another one of the methods outlined below focuses on the importance of features according to the relative contribution of each neuron to this feature.

\subsubsection{Local Interpretable Model-agnostic Explanations (LIME)}
Local interpretable model-agnostic explanations \cite{ribeiro2016should} --- or LIME for short --- is an algorithm that explains a complex model by using a linear approximation trained on the same data. The explanations used in LIME are calculated as below: 

\begin{equation}
\xi(x)=\arg\min_{g\in{}G}L(f,g,\pi_x)+\Omega(g)
\end{equation}

We have that:
\begin{enumerate}
    \item $f$ is the model to be explained --- and $f(x)$ is the probability of $x$ belonging to a class, or the probability of $x$ to belong to each class separately in a multilabel problem---;
    \item $g \in G$ is an interpretable classifier that belongs to a class of interpretable classifiers formally defined as $G$;
    \item $\Omega(g)$ is the measure of the complexity of the interpretable solution (e.g. a deep decision tree would be very complex, and thus would have a high $\Omega(g)$); and,
    \item $\pi$ is the measure of local fidelity, which is defined as the faithfulness of the interpretable model to the original model in the area of the data point that we are predicting on.
\end{enumerate}

LIME also provides an open-source Python package \cite{lime_github} that facilitates the use of this method to explain black-box models. This Python library includes extraction of the top input features as well as their visualization. Following the examples provided on the library's repository, we use a document extracted from the "20 newsgroups" dataset \cite{Lang95} to showcase LIME's different explainability capabilities.

For instance, getting top features from a document in the "20 newsgroups" for a document would result in: 
\begin{verbatim}
    ('Celebrate', 0.0037133567817147738)
    ('Assumption', 0.0036240994429636965)
    ('Orthodox', 0.00359829342658496)
\end{verbatim}

The LIME library also includes visualizations of the results, such as the probability distribution of the output label:
\begin{figure}[ht]
    \vskip 0.2in
    \begin{center}
    \centerline{\includegraphics[width=4cm]{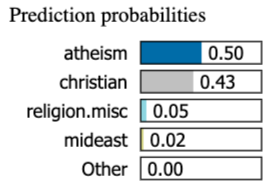}}
    \caption{An example of Probability Visualization from LIME}
    \label{fig:probs}
    \end{center}
    \vskip -0.2in
\end{figure}

Finally, LIME also provides a visualization of the original text with the most important terms highlighted in order to show the words influencing the prediction of a a class. 
\begin{figure}[ht]
    \vskip 0.2in
    \begin{center}
    \centerline{\includegraphics[width=\columnwidth]{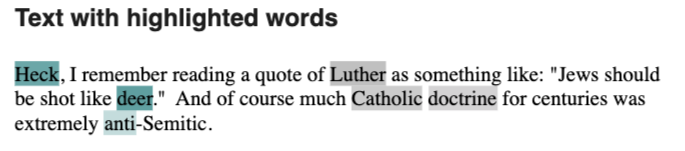}}
    \caption{An example of Word Highlighting Visualization from LIME}
    \label{fig:highlighted}
    \end{center}
    \vskip -0.2in
\end{figure}

\subsubsection{Deep Learning Important FeaTures (DeepLift)}
Deep Learning Important FeaTures \cite{shrikumar2017learning} --- or DeepLift for short --- is an explainability method based on an intuitive approach: the difference between the reference activation of a neuron and its activation with a modified input feature provides a good estimate of its contribution into the overall neural network. Let us take the example of a simple ReLU function for the sake of illustration. When the input of the function is $1$ ---i.e. $x = 1$---, the output will be $ReLU(1) = 1$. If we change the input to $x' = -1$, the output will be $ReLU(x') = 0$. However as $(\forall z<0): ReLu(z)=0$, this information is not significant.

To solve this inconsistency and focus on the infinitesimal input change from the baseline driving an output change from the baseline, DeepLift integrated gradients \cite{integrated_grad}. To calculate the gradient through the chain rule ---or \textit{slope} as defined in \cite{shrikumar2017learning}---, DeepLift redefines the gradient as $grad(F)=\frac{\Delta F}{\Delta x}$ for function $F$ and a given input $x$. Using this method, we can compute the contribution scores in a single backpropagation through the network.

\subsubsection{SHapley Additive exPlanation (SHAP)}
SHapley Additive exPlanation \cite{lundberg2018consistent} --- or SHAP for short ---  is an approach based on Shapley values used to determine feature attributions in game theory. Much like LIME, SHAP belongs to a class of explanation methods defined as \textit{additive feature attribution methods} as they are of the form:

\begin{equation}
g(z')=\phi_0+\sum_{j=1}^M\phi_jz_j' 
\end{equation}

We have that:
\begin{enumerate}
    \item $g$ is the model to be explained;
    \item $M$ is the number of input features;
    \item $z'\in\{0,1\}^M$ is the coalition vector (i.e. the vector presenting all observed features); and,
    \item $\phi_j\in\mathbb{R}$ is the weight of contributing feature $j$.
\end{enumerate}

This method builds a local approximation model as a linear combination of original features with an arbitrary input value. While LIME builds the local model by fitting a linear regression, SHAP computes the weight of each feature by leveraging the Shapley values ---which gives its name to this method. As outlined in the paper, the feature attributions generated by SHAP are "consistent and locally accurate attribution values" \cite{lundberg2018consistent}.

On top of outputting a series of feature attributions, SHAP also allows to visualize these features. Below is an example visualization of the feature attributions computed through SHAP of the document previously used to view the results provided by LIME:
\begin{figure}[ht]
    \vskip 0.2in
    \begin{center}
    \centerline{\includegraphics[width=\columnwidth]{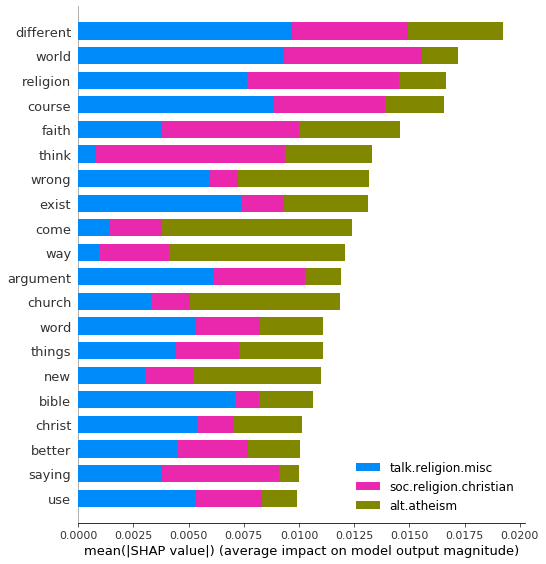}}
    \caption{An example of Probability Visualization from SHAP}
    \label{fig:SHAP}
    \end{center}
    \vskip -0.2in
\end{figure}

We see that the word “\textit{different}” is the most significant signal word used by the SHAP model, contributing most to the class \textit{misc (miscellaneous)} predictions. 
The word “\textit{religion}” on the other is the third biggest signal word used by the SHAP model, and contributes mostly to the classes \textit{misc} and \textit{christian}. It also tends to have a negative signal for all the other class as it is unlikely to see the word “\textit{religion}” appearing in a document about \textit{atheism} for example.

\section{Tentative Taxonomy}

Following the above outline of the most widely-used methods for explainability, we can summarize them into three main categories which span across the whole cycle of modelling:
\begin{enumerate}
    \item \textbf{Pre-modelling explainability}, which focuses on the study of the input;
    \item \textbf{Modelling explainability}, which focuses on the inner workings of the model – especially its mathematical aspect; and,
    \item \textbf{Post-modelling explainability}, which focuses on model approximation and results reporting.
\end{enumerate}

\subsection{Pre-modelling Explainability: Data-Dependent Methods}
The most rudimentary explainability method is data description and data analysis. Studying the input by plotting the data distribution, analysing the different classes, and even showing word tokens and class labels in the case of NLP, can all be used to explain the input data. Even though data exploration does not directly explain a model, it helps the user to understand some of the model's behaviors. For instance, knowing that the dataset is imbalanced can help understand why the classification performance is poor on some ---underrepresented--- labels.

Pre-modelling explainability also takes into account methods that rely on data perturbation to explain the inner workings of a model. DeepLift is one of such methods which compares the contribution of neurons between a baseline and the data point we want to explain to determine which features are the most significant in the final prediction.

\subsection{Modelling Explainability: Model Explanation Methods}
As the name suggests, model explanation methods rely on the model itself --- rather than relying on the data --- to provide understanding to the end-user. Such methods are inherently explainable models themselves, which are self-explanatory thanks to the presence of a human-fathomable number of explicitly defined parameters. Because DL models are black-boxes, they can only be explained through their sensitivity to the input (pre-modelling explainability) or through an approximation of their behavior (post-modelling explainability).

\subsection{Post-Modelling Explainability: Approximation Methods}
Approximation methods rely on the construction of proxy models which approximate the local behavior of a black-box model for a given input space. The main property of such proxy models is their inherent explainability as well as their faithfulness to the original model to be able to extrapolate its behavior. LIME and SHAP belong to this class of models which are used to explain a model after it was built and trained.

\section{Future Works}

This paper aims to introduce a new perspective on the explainability field by explaining its necessity in the industry and not only in research. However, this paper only scratches the surface of this fast-growing field. Some of the interesting interesting that can follow this article include:
\begin{itemize}
    \item a more comprehensive taxonomy allowing to have an established standard to refer to;
    \item a standardisation of the explainability framework, which is critical for researchers; and,
    \item an analysis of the explainability to complexity ratio of each explainability method, as well as the possible links between performance, complexity and explainability of widely used ML models.
\end{itemize}

As we have seen throughout this paper, there are multiple ways to explain ML models and algorithms; nonetheless, there is currently no standardised method to measure models' explainability. Indeed, only human judges can say if a model is explainable or not as the target of XAI is then ---human--- end-user. Relying only on this opinion is too subjective as it can vary widely from person to person. A model that is explainable for a user might not be explainable to another person. 

Another reason why explainability standardisation is paramount is industry-grade AI, which needs to be safe for the end-users. Standards could include questions such as: is this model controllable, is this model auditable, how interpretable is this model, how much is it explainable, ... This is very useful for the developers who will implement such models either by training them by themselves or by creating MaaS (Model as a Service) models.

\section{Conclusion}
Through this paper, we have defined the meaning of AI explainability and established its importance for the construction of trust-worthy and intentional Machine Learning models. We have also introduced the most widely-used explainability methods to contextualize this explanation and offer an overview to data scientists in the industry. This opens the door to a wider adoption of XAI and a systematization of this field; for if we do not drive explainability today, it will be too late tomorrow to correct the wrongs of AI and ensure a safe future where AI is omnipresent.


\newpage

\bibliography{bibliography}
\bibliographystyle{icml2020}

\end{document}